\documentclass{article} % For LaTeX2e
\usepackage{iclr2024_conference,times}

\usepackage{amsmath,amsfonts,bm}

\def\eqref#1{equation~\ref{#1}}
\def\1{\bm{1}}

\DeclareMathAlphabet{\mathsfit}{\encodingdefault}{\sfdefault}{m}{sl}
\SetMathAlphabet{\mathsfit}{bold}{\encodingdefault}{\sfdefault}{bx}{n}

\usepackage{bm}
\usepackage{graphicx}
\usepackage{xcolor}
\usepackage{amstext}
\usepackage{amsmath}
\usepackage{amsthm}
\usepackage{amssymb}
\usepackage{hyperref}
\usepackage{url}
\usepackage{algorithm}
\usepackage{algpseudocode}

\usepackage{thmtools}
\usepackage{thm-restate}
\usepackage{array, multirow, multicol}
\newtheorem{theorem}{Theorem}

\newtheorem{prop}{Proposition}

\algnewcommand\algorithmicinput{\textbf{Input:}}
\algnewcommand\Input{\item[\algorithmicinput]}
\algnewcommand\algorithmicoutput{\textbf{Output:}}
\algnewcommand\Output{\item[\algorithmicoutput]}
\newtheorem{pot}{Proof}
\title{Analytical Solution of a Three-layer Network with a Matrix Exponential Activation Function}

\author{Kuo Gai, Shihua Zhang \\
Academy of Mathematics and Systems Science\\
Chinese Academy of Sciences\\
Beijing, 100190, China \\
School of Mathematics Sciences\\
University of Chinese Academy of Science\\
Beijing, 100049, China \\
\texttt{\{gaikuo, zsh\}@amss.ac.cn} \\
}

\iclrfinalcopy % Uncomment for camera-ready version, but NOT for submission.
\begin{document}

\maketitle

\begin{abstract}
In practice, deeper networks tend to be more powerful than shallow ones, but this has not been understood theoretically. In this paper, we find a analytical solution of a three-layer network with a matrix exponential activation function, i.e., 
$$
\bm{f}(\bm{X})=\bm{W}_3\exp(\bm{W}_2\exp(\bm{W}_1\bm{X})), \bm{X}\in \mathbb{C}^{d\times d}
$$
have analytical solutions for the equations
$$
\left\{\begin{array}{c}
	\bm{Y}_1=\bm{f}(\bm{X}_1)  \\
	\bm{Y}_2=\bm{f}(\bm{X}_2) 
\end{array}\right.
$$
for $\bm{X}_1,\bm{X}_2,\bm{Y}_1,\bm{Y}_2$ with only invertible assumptions. Our proof shows the power of depth and the use of a non-linear activation function, since one layer network can only solve one equation,i.e.,$\bm{Y}=\bm{W}\bm{X}$.
\end{abstract}

\section{Introduction}
Deep neural networks have become successful in many fields, including computer vision, natural language processing, bioinformatics, etc. However, the mathematical principle of deep learning is still not fully understood, especially why deeper networks with non-linear activation functions tend to be more powerful than shallower ones.

It is well known that sufficient large depth-2 neural networks with reasonable activation functions can approximate any continuous function on a bounded domain \citep{cybenko1989approximation,funahashi1989approximate,hornik1989multilayer,barron1994approximation,pinkus1999approximation}, but this requires the width of networks to be exponential. Recent authors have shown that some functions can be approximated by deeper networks with fewer neurons than by shallower ones, such as radial functions \citep{eldan2016power}, Boolean circuit \citep{rossman2015average} or functions induced by neural network \citep{telgarsky2016benefits}. However, these functions are far from the function approximated by neural networks in practice.

There are also some studies on approximating data points of a fixed number instead of continuous functions, which is more general since data points can be sampled from arbitrary distributions. However, such works focus more on width rather than depth. For instance, the notable framework neural tangent kernel(NTK)\citep{jacot2018neural} proved that neural networks can fit the data with error 0 if the width is infinite. However, such wide neural networks would also have an extremely large number of parameters, and extract random features of data. Moreover, current state of art results are typically achieved by deep neural networks \citep{he2016deep,krizhevsky2012imagenet}. Generally, when the width of the network is bounded since the function class of neural networks becomes more complex after the composition of layers, the optimization process of neural networks 
may not find the global optimal solution. There are some empirical explorations which reveal non-trivial properties of the landscape \citep{goodfellow2014qualitatively,li2018visualizing}. However, these properties still lack theoretical understanding since the optimization of network is highly non-convex. Thus, to show the power of depth, a potential way is to pursue analytical solution instead of optimization. A line of research focuses on memory capacity \citep{vershynin2020memory,yamasaki1993lower,huang2003learning,zhang2021understanding,yun2019small}, which aims at proving the existence of solutions through construction rather than computation. The construction is tricky and the labels are limited to be scalars.

Some studies are using the matrix-form activation function in practice. \citet{li2017second} introduces the use of a matrix operation (either matrix logarithm or matrix square root) on top of a convolutional layer with higher-order feature crosses. \citep{fischbacher2020intelligent} proposes a single matrix exponential layer to learn the periodic structure or geometric invariants of the input. Matrix-form activation functions make it possible to find the solution through matrix computation instead of construction and provide a better understanding of the power of depth and non-linear activation functions.

In this paper, we omit the optimization process and compute the analytical solution of a three-layer neural network with a matrix exponential activation function. We show the power of depth by proving that a three-layer network can map more matrix-form data points to their labels than a single-layer network. We also shed light on networks with element-wise activation function experimentally using similar methodology, indicating the number of equations a network can solve increases with the number of layers linearly. 

\section{Preliminary}\label{sec:pre}
The matrix exponential is a matrix function on the square matrices analogous to the ordinary exponential function. Let $\bm{X}$ be an $d\times d$ complex matrix. The exponential of $\bm{X}$, denoted by $\operatorname{exp}(\bm{X})$ is the $d\times d$ matrix given by the power series
\begin{equation}
	\operatorname{exp}(\bm{X})=\sum_{k=0}^{\infty}\frac{1}{k!}\bm{{X}}^k
\end{equation}
where $\bm{X}^{0}$ is defined to be the identity matrix $\bm{I}$ with the same dimensions as $\bm{X}$. The matrix exponential is well studied in the theory of Lie group and has many good properties.
\begin{prop}\label{prop1}
	Let $\bm{X},\bm{Y}\in \mathbb{C}^{d\times d}$. If $\bm{X}\bm{Y}=\bm{Y}\bm{X}$, then $\operatorname{exp}(\bm{X})\operatorname{exp}(\bm{Y})=\operatorname{exp}(\bm{X}+\bm{Y})$ 
\end{prop}
\begin{prop}\label{prop2}
	The matrix exponential gives a surjective map 
	\begin{equation}
		\operatorname{exp}:M_d(\mathbb{C})\to \operatorname{GL}(d,\mathbb{C})
	\end{equation}
	where $M_d(\mathbb{C})$ is the space of all $d\times d$ complex matrices and $\operatorname{GL}(d,\mathbb{C})$ is the general linear group of degree $d$, i.e. the group of all $d\times d$ invertible matrices.
\end{prop}
In general, $\operatorname{exp}(\bm{X})\operatorname{exp}(\bm{Y})$ can be expressed by the Baker Campbell Hausdorff (BCH) formula, and when $\bm{X}$ and $\bm{Y}$ commute, the computation of BCH formula can be simplified as in Proposition \ref{prop1}. Proposition \ref{prop2} means every invertible matrix $\bm{X}$ can be written as the exponential of some other matrix $\bm{Z}$ (for this, it is essential to consider the field $\mathbb{C}$ and not $\mathbb{R}$). 

$\bm{Z}$ can be calculated through the logarithm of matrix. First we need to find the Jordan decomposition of $X$ and calculate the logarithm of the Jordan blocks. For instance, we can write a Jordan block as 
\begin{equation*}
	\begin{aligned}
		\bm{B}&=\left[\begin{array}{cccccc}
			\lambda & 1 & 0 & 0 & \cdots & 0 \\
			0 & \lambda & 1 & 0 & \cdots & 0 \\
			0 & 0 & \lambda & 1 & \cdots & 0 \\
			\vdots & \vdots & \vdots & \ddots & \ddots & \vdots \\
			0 & 0 & 0 & 0 & \lambda & 1 \\
			0 & 0 & 0 & 0 & 0 & \lambda
		\end{array}\right]\\
	\end{aligned}
\end{equation*}
\begin{equation}
	\begin{aligned}
		&=\lambda\left[\begin{array}{cccccc}
			1 & \lambda^{-1} & 0 & 0 & \cdots & 0 \\
			0 & 1 & \lambda^{-1} & 0 & \cdots & 0 \\
			0 & 0 & 1 & \lambda^{-1} & \cdots & 0 \\
			\vdots & \vdots & \vdots & \ddots & \ddots & \vdots \\
			0 & 0 & 0 & 0 & 1 & \lambda^{-1} \\
			0 & 0 & 0 & 0 & 0 & 1
		\end{array}\right]\\
		&=\lambda(\bm{I}+\bm{K})
	\end{aligned}
\end{equation}
where $\bm{K}$ is a matrix with zeros on and under the main diagonal. The number $\lambda$ is nonzero by the assumption that $\bm{X}$ is invertible. Then, by the Mercator series
\begin{equation}
	\log(1+x)=x-\frac{x^2}{2}+\frac{x^3}{3}-\frac{x^4}{4}+\cdots    
\end{equation}
we have 
\begin{equation}
	\log \bm{B}=\log(\lambda(\bm{I}+\bm{K}))=(\log \lambda)\bm{I}+\bm{K}-\frac{\bm{K}^2}{2}+\frac{\bm{K}^3}{3}-\frac{\bm{K}^4}{4}+\cdots
\end{equation}
This series has a finite number of terms since $\bm{K}^m$ is $\bm{0}$ if $m$ is the dimension of of $\bm{K}$. Thus the sum is well-defined. Assume that $\bm{J}$ is the Jordan normal form of $\bm{X}$ and $\bm{X}=\bm{P}\bm{J}\bm{P}^{-1}$. Following the method above, we can calculate $\log\bm{J}$ and obtain $\bm{Z}=\log\bm{X}=\bm{P}\log\bm{J}\bm{P}^{-1}$. 
\section{Main result}
The basic task of machine learning is to find a function which maps the data to its label, i.e., for given $\{(\bm{x}_i,\bm{y}_i)\}_{i=1}^n$ where $\bm{x}_i\in \mathbb{R}^{d_x}, \bm{y}_i\in \mathbb{R}^{d_y}$, solve the equations $f(\bm{x}_i)=\bm{y}_i$, $i=1,\cdots,n$. Specifically, for neural networks, $f$ is composed of linear transformations and nonlinear activation functions, i.e., for $m$-layer network,
\begin{equation}
	\bm{f}(\cdot)=\bm{W}_m\sigma\left(\bm{W}_{m-1}\cdots \sigma \left(\bm{W}_{1}\cdot\right)\right)
\end{equation}
where $\sigma$ is the nonlinear activation function and $\bm{W}_{1}\in \mathbb{R}^{d_x\times d_1}$, $\bm{W}_{k}\in \mathbb{R}^{d_{k-1}\times d_{k}}$, $k=2,\cdots,m-1$, $\bm{W}_{m}\in \mathbb{R}^{d_{m-1}\times d_y}$. $\sigma$ is elementwise function such as ReLU, sigmoid and tanh function. Generally, proving the existence of solution of nonlinear system is hard, especially when the element-wise function $\sigma$ does not integral well with the linear transformation matrix $\bm{W}$. For instance, let $\sigma(x)=x^2$, then $\sigma(\bm{A})=\bm{A}\circ \bm{A}$ for $\bm{A}\in \mathbb{R}^{d\times d^{\prime}}$, where $\circ$ is the Hadamard product. As we know, generally, $\bm{A}\circ \bm{A}$ can not be expressed as a polynomial of $\bm{A}$, i.e., $\bm{A}\circ \bm{A}\neq \operatorname{poly}(\bm{A})$. This causes difficulties in finding the analytical solution of neural networks, since we can not transform the output of each layer to a operable form. To address this issue, we use matrix exponential function as nonlinear activation function instead, which gives chance to find the solution to the system when number of layers is more than one.  

To make matrix exponential well-defined, we assume $\bm{X},\bm{Y},\bm{W}$ are square. To make the solution exists, we assume the items of $\bm{X},\bm{Y},\bm{W}$ in $\mathbb{C}$. Consider $\bm{X}, \bm{Y}\in \mathbb{C}^{d\times d}$ and $\bm{X}$ is invertible, then $\bm{W}=\bm{Y}\bm{X}^{-1}$ can solve the equation $\bm{Y}=\bm{W}\bm{X}$. There doesn't exist solution of $\bm{Y}_1=\bm{W}\bm{X}_1, \bm{Y}_2=\bm{W}\bm{X}_2$ for $\bm{X}_1,\bm{X}_2,\bm{Y}_1,\bm{Y}_2\in \mathbb{C}^{d\times d}$ except degenerate cases, since the number of parameter $d^2$ is less than the number of equations $2d^2$. If we let the weight matrix be `wider', i.e., 
\begin{equation}
	\bm{W}=\left[\begin{array}{cc}
		\bm{W}_1 & \bm{0} \\
		\bm{0} & \bm{W}_2
	\end{array}\right]
\end{equation}
then with the assumption that $\bm{X}_1$ and $\bm{X}_2$ are invertible, $\bm{W}_1=\bm{Y}_1\bm{X}_1^{-1}$ and $\bm{W}_2=\bm{Y}_2\bm{X}_2^{-1}$ can solve the equations
\begin{equation}
	\left[\begin{array}{c}
		\bm{Y}_1 \\
		\bm{Y}_2
	\end{array}\right]=\left[\begin{array}{cc}
		\bm{W}_1 & \bm{0} \\
		\bm{0} & \bm{W}_2
	\end{array}\right]\cdot\left[\begin{array}{c}
		\bm{X}_1 \\
		\bm{X}_2
	\end{array}\right]
\end{equation}
The above equation has solution because we can separate it to two sub-problems and solve $\bm{W}_1$ and $\bm{W}_2$ sequentially. However, this will not happen when we compose $\bm{W}_1$ and $\bm{W}_2$ (two-layer network with identity activation function), which means, solving the equation
\begin{equation}
	\bm{Y}_1=\bm{W}_2\bm{W}_1\bm{X}_1;\bm{Y}_2=\bm{W}_2\bm{W}_1\bm{X}_2
\end{equation}
When $\bm{W}_1$ is fixed, then $\bm{W}_2$ with $d^2$ parameters is involved in $2d^2$ equations, i.e., $\bm{Y}_1=\bm{W}_2(\bm{W}_1\bm{X}_1)$ and $\bm{Y}_2=\bm{W}_2(\bm{W}_1\bm{X}_2)$ and has no solution in general.
Situation changes again by adding non-linear activation function, i.e., solving the equations
\begin{equation}\label{eq:nlsys}
	\begin{aligned}
		&\bm{Y}_1=\bm{W}_2\sigma(\bm{W}_1\bm{X}_1)\\
		&\bm{Y}_2=\bm{W}_2\sigma(\bm{W}_1\bm{X}_2)
	\end{aligned}
\end{equation}
From the second equation, we obtain $\bm{W}_2=\bm{Y}_2\sigma(\bm{W}_1\bm{X}_2)^{-1}$. Taking it into the first equation, we have
\begin{equation}\label{eq:solvetwo}
	\bm{Y}_1=\bm{Y}_2\sigma(\bm{W}_1\bm{X}_2)^{-1}\sigma(\bm{W}_1\bm{X}_1)
\end{equation}
If this equation has a solution for $\bm{W}_1$, then the non-linear system (\ref{eq:nlsys}) has a solution for $\bm{W}_1$ and $\bm{W}_2$. Following this intuition, we prove that a three-layer network with a matrix exponential activation function can solve the equations, exhibiting the power of deepness and the use of non-linear activation.

\begin{theorem}\label{thm:1}
	Let $\bm{X}_1,\bm{X}_2$ be the data matrices and $\bm{Y}_1,\bm{Y}_2$ be the corresponding label matrices, where $\bm{X}_1,\bm{X}_2,\bm{Y}_1,\bm{Y}_2\in \mathbb{C}^{d\times d}$ are invertible matrices. Assume that $\bm{X}_1-\bm{X}_2$ is invertible. $\bm{f}(\cdot)=\bm{W}_3\sigma(\bm{W}_2\sigma(\bm{W}_1\cdot))$ is a three-layer network where $\sigma(\cdot)$ is matrix exponential, i.e., $\sigma(\cdot)=\operatorname{exp}(\cdot):\mathbb{C}^{d\times d}\to \mathbb{C}^{d\times d}$, and $\bm{W}_1,\bm{W}_2,\bm{W}_3\in \mathbb{C}^{d\times d}$. If 
	\begin{equation}
		\begin{aligned}
			\bm{W}_1&=\operatorname{ln}\alpha \cdot (\bm{X}_1-\bm{X}_2)^{-1}\\
			\bm{W}_2&=(\bm{Z}-\operatorname{ln}\alpha\cdot \bm{I})\cdot \operatorname{exp}(-\bm{W}_1\bm{X}_2)\cdot \frac{1}{1-\alpha}\\
			\bm{W}_3&=\bm{Y}_1\operatorname{exp}(-\bm{W}_2 \operatorname{exp}(\bm{W}_1\bm{X}_1))
		\end{aligned}
	\end{equation}
	
	where $\alpha\in \mathbb{R}^{+}, \alpha \neq 1$ and $\operatorname{exp}(\bm{Z})=\alpha \bm{Y}_1^{-1}\bm{Y}_2$, then $\bm{f}$ maps the data points to their labels, i.e., $\bm{f}(\bm{X}_1)=\bm{Y}_1, \bm{f}(\bm{X}_2)=\bm{Y}_2$
\end{theorem}

\begin{pot}
	We assume 
	\begin{align}
		\bm{W}_1=\bm{W}_{1,1}\bm{W}_{1,2}
	\end{align}
	where $\bm{W}_{1,1},\bm{W}_{1,2}\in \mathbb{C}^{d\times d}$ and $\bm{W}_{1,2}$ is invertible.
	It is known that the exponential of a matrix is always an invertible matrix, let
	\begin{equation}
		\begin{aligned}
			\bm{M}_{1,X_1}&=\operatorname{exp}(\bm{W}_1\bm{X}_1)\bm{X}_1^{-1}\bm{W}_{1,2}^{-1}\\[1mm]
			\bm{M}_{1,X_2}&=\operatorname{exp}(\bm{W}_1\bm{X}_2)\bm{X}_2^{-1}\bm{W}_{1,2}^{-1}\\[1mm]
			\bm{M}_{2,X_1}&=\operatorname{exp}(\bm{W}_2\operatorname{exp}(\bm{W}_1\bm{X}_1))\operatorname{exp}(\bm{W}_1\bm{X}_1)^{-1}\\[1mm]
			\bm{M}_{2,X_2}&=\operatorname{exp}(\bm{W}_2\operatorname{exp}(\bm{W}_1\bm{X}_2))\operatorname{exp}(\bm{W}_1\bm{X}_2)^{-1}\\[1mm]
		\end{aligned}
	\end{equation}
	
	Use the trick 
	\begin{equation}
		\left[\begin{array}{cc}
			\bm{A} & \bm{0} \\
			\bm{0} & \bm{B}
		\end{array}\right]=\left[\begin{array}{cc}
			\bm{A} & \bm{0} \\
			\bm{0} & \bm{A}
		\end{array}\right]\cdot\left[\begin{array}{cc}
			\bm{I} & \bm{0} \\
			\bm{0} & \bm{A}^{-1}\bm{B}
		\end{array}\right]
	\end{equation}
	twice, then we have 
	\begin{equation}\label{eq0}
		\begin{aligned}
			&\left[\begin{array}{cc}
				\operatorname{exp}(\bm{W}_2\operatorname{exp}(\bm{W}_1\bm{X}_1))&0\\
				0&\operatorname{exp}(\bm{W}_2\operatorname{exp}(\bm{W}_1\bm{X}_2))\\
			\end{array}\right]\\[2mm]
			=&\left[\begin{array}{cc}
				\bm{M}_{2,X_1}&0\\
				0&\bm{M}_{2,X_2}\\
			\end{array}\right]\cdot
			\left[\begin{array}{cc}
				\bm{M}_{1,X_1}&0\\
				0&\bm{M}_{1,X_2}\\
			\end{array}\right]\\
			\cdot&\left[\begin{array}{cc}
				\bm{W}_{1,2}\bm{X}_1&0\\
				0&\bm{W}_{1,2}\bm{X}_{2}\\
			\end{array}\right]\\[2mm]
		\end{aligned}
	\end{equation}
	\begin{equation*}
		\begin{aligned}
			=&\left[\begin{array}{cc}
				\bm{M}_{2,X_1}&0\\
				0&\bm{M}_{2,X_2}\\
			\end{array}\right]\cdot\left[\begin{array}{cc}
				\bm{M}_{1,X_2}&0\\
				0&\bm{M}_{1,X_2}\\
			\end{array}\right]\\
			\cdot&\left[\begin{array}{cc}
				\bm{M}_{1,X_2}^{-1}\bm{M}_{1,X_1}&0\\
				0&\bm{I}\\
			\end{array}\right]\cdot
			\left[\begin{array}{cc}
				\bm{W}_{1,2}\bm{X}_1&0\\
				0&\bm{W}_{1,2}\bm{X}_{2}\\
			\end{array}\right]\\[2mm]
			=&\left[\begin{array}{cc}
				\bm{M}_{2,X_1}\bm{M}_{1,X_2}&0\\
				0&\bm{M}_{2,X_2}\bm{M}_{1,X_2}\\
			\end{array}\right]\cdot
			\left[\begin{array}{cc}
				\bm{M}_{1,X_2}^{-1}\bm{M}_{1,X_1}&0\\
				0&\bm{I}\\
			\end{array}\right]\\
			\cdot&\left[\begin{array}{cc}
				\bm{W}_{1,2}\bm{X}_1&0\\
				0&\bm{W}_{1,2}\bm{X}_{2}\\
			\end{array}\right]\\[2mm]
			=&\left[\begin{array}{cc}
				\bm{M}_{2,X_1}\bm{M}_{1,X_2}&0\\
				0&\bm{M}_{2,X_1}\bm{M}_{1,X_2}\\
			\end{array}\right]\\
			\cdot&\left[\begin{array}{cc}
				\bm{I}&0\\
				0&\bm{M}_{1,X_2}^{-1}\bm{M}_{2,X_1}^{-1}\bm{M}_{2,X_2}\bm{M}_{1,X_2}\\
			\end{array}\right]\\
			\cdot&\left[\begin{array}{cc}
				\bm{M}_{1,X_2}^{-1}\bm{M}_{1,X_1}&0\\
				0&\bm{I}\\
			\end{array}\right]
			\cdot\left[\begin{array}{cc}
				\bm{W}_{1,2}\bm{X}_1&0\\
				0&\bm{W}_{1,2}\bm{X}_{2}\\
			\end{array}\right]\\	
		\end{aligned}
	\end{equation*}
	
	Let  
	\begin{equation}
		\label{eq7}
		\bm{W}_3=\bm{M}_{1,X_2}^{-1}\bm{M}_{2,X_1}^{-1},
	\end{equation}
	to eliminate the fist matrix of the right side of the last equality in (\ref{eq0}), then we have
	\begin{equation}
		\begin{aligned}
			&\left[\begin{array}{cc}
				\bm{f}(\bm{X}_1)&0\\
				0&\bm{f}(\bm{X}_2)\\
			\end{array}\right]\\
			=&\left[\begin{array}{cc}
				\bm{W}_3\operatorname{exp}(\bm{W}_2\operatorname{exp}(\bm{W}_1\bm{X}_1))&0\\
				0&\bm{W}_3\operatorname{exp}(\bm{W}_2\operatorname{exp}(\bm{W}_1\bm{X}_2))\\
			\end{array}\right]\\[2mm]
			=&\left[\begin{array}{cc}
				\bm{I}&0\\
				0&\bm{M}_{1,X_2}^{-1}\bm{M}_{2,X_1}^{-1}\bm{M}_{2,X_2}\bm{M}_{1,X_2}\\
			\end{array}\right]\cdot
			\left[\begin{array}{cc}
				\bm{M}_{1,X_2}^{-1}\bm{M}_{1,X_1}&0\\
				0&\bm{I}\\
			\end{array}\right]\\
			\cdot&
			\left[\begin{array}{cc}
				\bm{W}_{1,2}\bm{X}_1&0\\
				0&\bm{W}_{1,2}\bm{X}_{2}\\
			\end{array}\right]\\
		\end{aligned}
	\end{equation}
	
	Let $\bm{\tilde{X}}_1=\bm{W}_{1,2}\bm{X}_1, \bm{\tilde{X}}_2=\bm{W}_{1,2}\bm{X}_2$. To solve 
	\begin{equation}
		\left[\begin{array}{cc}
			\bm{f}(\bm{X}_1)&0\\
			0&\bm{f}(\bm{X}_2)\\
		\end{array}\right]=
		\left[\begin{array}{cc}
			\bm{Y}_1&0\\
			0&\bm{Y}_2\\
		\end{array}\right],
	\end{equation}
	it equals to solve 
	\begin{equation}\label{eq1}
		\left\{\begin{array}{c}
			\bm{M}_{1,X_2}^{-1}\bm{M}_{1,X_1}\bm{\tilde{X}}_1=\bm{Y}_1 \\[6mm]
			\bm{M}_{1,X_2}^{-1}\bm{M}_{2,X_1}^{-1}\bm{M}_{2,X_2}\bm{M}_{1,X_2}\bm{\tilde{X}}_2=\bm{Y}_2
		\end{array}\right.
	\end{equation}
	By the definition of $\bm{M}_{1,X_1},\bm{M}_{1,X_2},\bm{M}_{2,X_1},\bm{M}_{2,X_2}$, we can rewrite equalities in (\ref{eq1}) as:
	\begin{equation}\label{eq2}
		\begin{scriptsize}
			\left\{\begin{array}{l}
				\bm{\tilde{X}}_2\operatorname{exp}(\bm{W}_{1,1}\bm{\tilde{X}}_2)^{-1}  \operatorname{exp}(\bm{W}_{1,1}\bm{\tilde{X}}_1)=\bm{Y}_1\\[6mm]
				\bm{\tilde{X}}_2\operatorname{exp}(\bm{W}_{1,1}\bm{\tilde{X}}_2)^{-1}\operatorname{exp}(\bm{W}_{1,1}\bm{\tilde{X}}_1)\operatorname{exp}(\bm{W}_2\operatorname{exp}(\bm{W}_{1,1}\bm{\tilde{X}}_1))^{-1}\operatorname{exp}(\bm{W}_2\operatorname{exp}(\bm{W}_{1,1}\bm{\tilde{X}}_2))=\bm{Y}_2  
			\end{array}\right. 
		\end{scriptsize}
	\end{equation}
	To solve the first equality in (\ref{eq2}), let 
	\begin{equation}
		\bm{W}_{1,2}=\frac{1}{\alpha}\bm{Y}_1\bm{X}_2^{-1}
	\end{equation}
	where $\alpha\in \mathbb{R}^{+}, \alpha \neq 1$, then 
	\begin{equation}
		\bm{\tilde{X}}_2^{-1}\bm{Y}_1=\alpha \bm{I}=\operatorname{exp}(\operatorname{ln}\alpha \cdot \bm{I})
	\end{equation}
	Then the first equality in (\ref{eq2}) can be rewrite as 
	\begin{equation}
		\label{eq5}
		\begin{aligned}
			\operatorname{exp}(\bm{W}_{1,1}\bm{\tilde{X}}_1)&=\operatorname{exp}(\bm{W}_{1,1}\bm{\tilde{X}}_2)\operatorname{exp}(\operatorname{ln}\alpha \cdot \bm{I})\\[4mm]
			&=\operatorname{exp}(\bm{W}_{1,1}\bm{\tilde{X}}_2+\operatorname{ln}\alpha \cdot \bm{I})	
		\end{aligned}
	\end{equation}
	The second equality is because $\bm{W}_{1,1}\bm{\tilde{X}}_2$ commute with $\operatorname{ln}\alpha \cdot \bm{I}$ and Proposition \ref{prop1}. Thus it is sufficient to solve the equality 
	\begin{equation}
		\bm{W}_{1,1}\bm{\tilde{X}}_1=\bm{W}_{1,1}\bm{\tilde{X}}_2+\operatorname{ln}\alpha \cdot \bm{I}
	\end{equation}
	since $\bm{X}_1-\bm{X}_2$ is invertible as assumed, then 
	\begin{equation}\label{eq9}
		\bm{W}_{1,1}=\operatorname{ln}\alpha \cdot (\bm{\tilde{X}}_1-\bm{\tilde{X}}_2)^{-1},\quad \bm{W}_{1}=\bm{W}_{1,1}\bm{W}_{1,2}=\operatorname{ln}\alpha \cdot (\bm{X}_1-\bm{X}_2)^{-1}
	\end{equation}
	Taking the second equality in (\ref{eq2}) into the first equality in (\ref{eq2}), the first equality in (\ref{eq2}) can be rewrite as 
	\begin{equation}
		\begin{aligned}
			\operatorname{exp}(\bm{W}_2\operatorname{exp}(\bm{W}_{1,1}\bm{\tilde{X}}_1))^{-1}\operatorname{exp}(\bm{W}_2\operatorname{exp}(\bm{W}_{1,1}\bm{\tilde{X}}_2))&=\bm{Y}_1^{-1}\bm{Y}_2\\
			&=\frac{1}{\alpha} \operatorname{exp}(\bm{Z}) 
		\end{aligned}
	\end{equation}
	The second equality is because of the definition of $\bm{Z}$. Such $\bm{Z}$ exists because of Proposition \ref{prop2}. If $\bm{W}_2\operatorname{exp}(\bm{W}_{1,1}\bm{\tilde{X}}_1)$ commute with $\bm{Z}$, then we only need to solve 
	\begin{equation}
		\label{eq6}
		\bm{W}_2\operatorname{exp}(\bm{W}_{1,1}\bm{\tilde{X}}_2)=\bm{W}_2\operatorname{exp}(\bm{W}_{1,1}\bm{\tilde{X}}_1)+(\bm{Z}-\operatorname{ln}\alpha\cdot \bm{I})
	\end{equation}
	Note that according to (\ref{eq5})
	\begin{equation}
		\begin{aligned}
			\operatorname{exp}(\bm{W}_{1,1}\bm{\tilde{X}}_2)-\operatorname{exp}(\bm{W}_{1,1}\bm{\tilde{X}}_1)&=\operatorname{exp}(\bm{W}_{1,1}\bm{\tilde{X}}_2)(\bm{I}-\alpha\bm{I})\\
			&=(1-\alpha)\operatorname{exp}(\bm{W}_{1,1}\bm{\tilde{X}}_2)
		\end{aligned}
	\end{equation}
	then $\operatorname{exp}(\bm{W}_{1,1}\bm{\tilde{X}}_2)-\operatorname{exp}(\bm{W}_{1,1}\bm{\tilde{X}}_1)$ is invertible since $\alpha\neq 1$. Thus the solution to (\ref{eq6}) is
	\begin{equation}
		\label{eq8}
		\begin{aligned}
			\bm{W}_2&=(\bm{Z}-\operatorname{ln}\alpha\cdot \bm{I})(\operatorname{exp}(\bm{W}_{1,1}\bm{\tilde{X}}_2)-\operatorname{exp}(\bm{W}_{1,1}\bm{\tilde{X}}_1))^{-1}\\[2mm]
			&=\frac{1}{1-\alpha}(\bm{Z}-\operatorname{ln}\alpha\cdot \bm{I})\operatorname{exp}(\bm{W}_{1,1}\bm{\tilde{X}}_2)^{-1}
		\end{aligned}
	\end{equation}
	Finally we need to verify that $\bm{W}_2\operatorname{exp}(\bm{W}_{1,1}\bm{\tilde{X}}_1)$ commute with $\bm{Z}$, it is obviously according to (\ref{eq5}) since 
	\begin{equation}
		\begin{small}
			\begin{aligned}
				\bm{W}_2\operatorname{exp}(\bm{W}_{1,1}\bm{\tilde{X}}_1)&=\frac{1}{1-\alpha}(\bm{Z}-\operatorname{ln}\alpha\cdot \bm{I})\operatorname{exp}(\bm{W}_{1,1}\bm{\tilde{X}}_2)^{-1}\operatorname{exp}(\bm{W}_{1,1}\bm{\tilde{X}}_1)\\
				&=\frac{\alpha}{1-\alpha}(\bm{Z}-\operatorname{ln}\alpha\cdot \bm{I})
			\end{aligned}	
		\end{small}
	\end{equation}
	When $\bm{W}_1, \bm{W}_2$ are fixed as (\ref{eq9}) and (\ref{eq8}), then $\bm{W}_3$ is fixed
	\begin{equation}
		\begin{aligned}
			\bm{W}_3&=\bm{M}_{1,X_2}^{-1}\bm{M}_{2,X_1}^{-1}\\[2mm]
			&=\bm{Y}_1\operatorname{exp}(-\bm{W}_2 \operatorname{exp}(\bm{W}_1\bm{X}_1))
		\end{aligned}
	\end{equation} 
	which concludes the proof.
\end{pot}

Note that $\bm{Z}$ can be calculated using the method in Section \ref{sec:pre}, thus the solution given in Theorem 1 can be calculated without gradient descent. The only assumption of data is $\bm{X}_1-\bm{X}_2$ is invertible, which is much more general than a certain class of functions.

\section{Experimental Results}\label{sec:exp}
Since we already found the analytical solution of a three-layer network with matrix exponential activation function, numerical experiments is not necessary. In this section, we focus on experiments on element-wise activation functions such as Relu and sigmoid using similar method. As discussed in Section \ref{sec:pre}, similar equation for two-layer network with element-wise activation $\sigma$, i.e.,
\begin{equation}
	\left\{\begin{array}{c}
		\bm{Y}_1=\bm{W}_2\sigma (\bm{W}_1\bm{X}_1)\\
		\bm{Y}_2=\bm{W}_2\sigma (\bm{W}_1\bm{X}_2)
	\end{array}\right.
\end{equation}
which equals to solving $\bm{W}_1$ and $\bm{W}_2$ sequentially through 
\begin{equation}\label{two-layer}
	\left\{\begin{array}{c}
		\bm{Y}_1=\bm{Y}_2\sigma(\bm{W}_1\bm{X}_2)^{-1}\sigma(\bm{W}_1\bm{X}_1)\\
		\bm{W}_2=\bm{Y}_2\sigma(\bm{W}_1\bm{X}_2)^{-1}
	\end{array}\right.
\end{equation}
In our experiments, we optimize $\|\bm{Y}_1-\bm{Y}_2\sigma(\bm{W}_1\bm{X}_2)^{-1}\sigma(\bm{W}_1\bm{X}_1)\|_F^2$ with gradient descent. Each item of $\bm{X}_1$,$\bm{X}_2$, $\bm{Y}_1$ and $\bm{Y}_2$ is sampled from Gaussian distribution $\mathcal{N}(0,1)$. For comparison, we compute the same value when $\sigma$ is the identity function, i.e., $\|\bm{Y}_1-\bm{Y}_2(\bm{W}_1\bm{X}_2)^{-1}\bm{W}_1\bm{X}_1\|_F^2=\|\bm{Y}_1-\bm{Y}_2\bm{X}_2^{-1}\bm{X}_1\|_F^2$. Then we can construct a score to measure the benefit of using sigmoid function or ReLU function in the training process
\begin{equation}
	s=\frac{\|\bm{Y}_1-\bm{Y}_2\sigma(\bm{W}_1\bm{X}_2)^{-1}\sigma(\bm{W}_1\bm{X}_1)\|_F^2}{\|\bm{Y}_1-\bm{Y}_2\bm{X}_2^{-1}\bm{X}_1\|_F^2}
\end{equation}

\begin{figure*}[h]
	\centering
	\includegraphics[width=0.99\columnwidth]{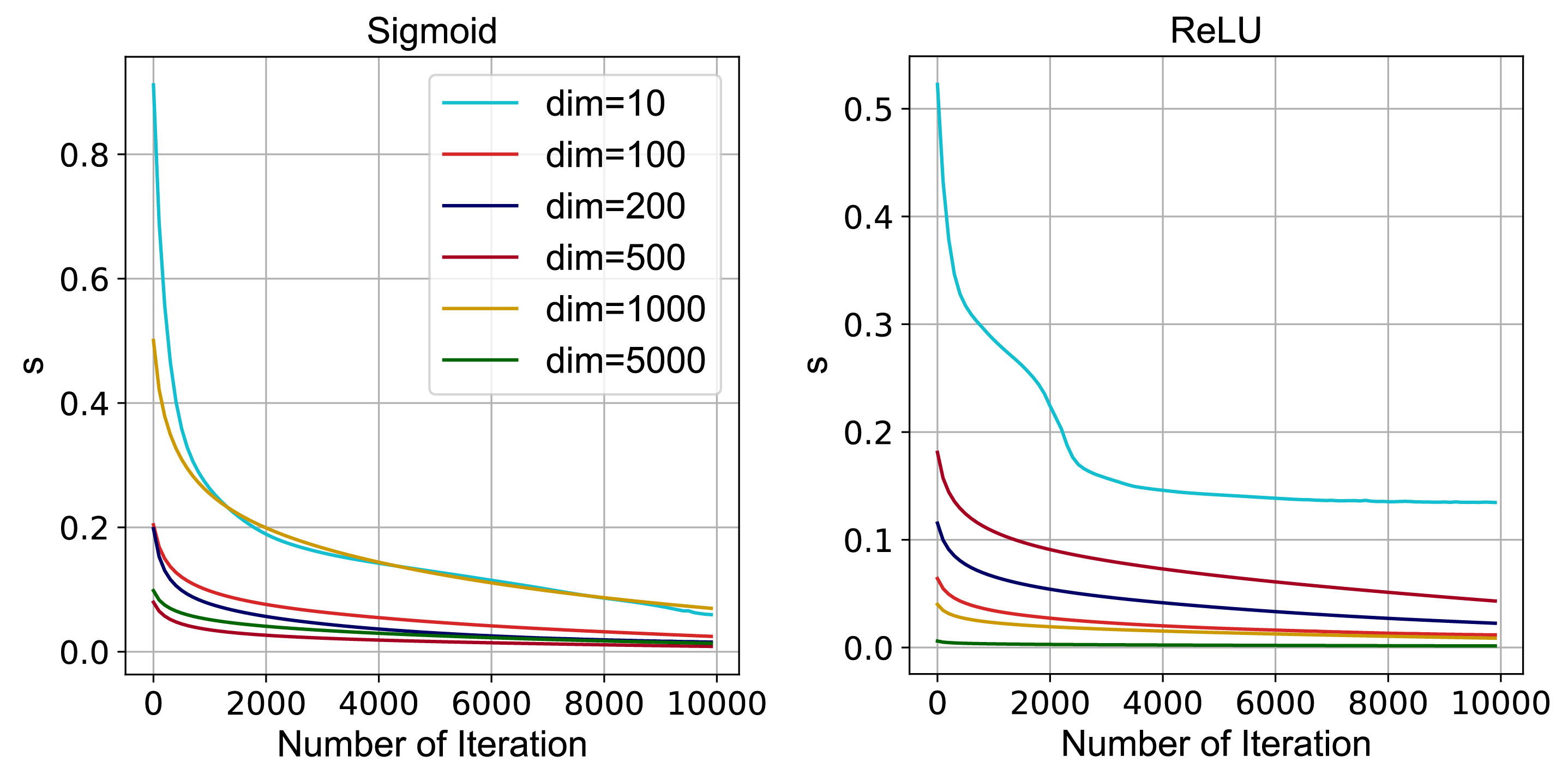}
	\caption{The $s$ score of two-layer network with Sigmoid (left) and ReLU (right) activation function in the training process.}
	\label{fig:2}
\end{figure*}

In the experiment (Fig.\ref{fig:2}), we find that both ReLU and Sigmoid function can find the optimal $\bm{W}_1$ with $s$ close to 0. This indicates that a two-layer network with ReLU or Sigmoid activation function has obvious benefits compared with the identity function and has the potential to solve twice the number of equations. Also the $s$ score decrease with the increasing of dimension, which means, the optimization problem becomes easier in high dimention space. However, it is hard to prove the existence of a solution of equality (\ref{two-layer}) and the existence of a path from initial weights to global optimal weights with gradient descent.

\section{Conclusion}\label{sec:discussion}
In this paper, we design a problem for a three-layer network with matrix exponential as an activation function and find the analytical solution. By doing this, we show the power of depth by comparing our three-layer networks to single-layer ones. Our result has merit compared with existing studies, both the studies finding special functions to show the power of depth and studies analyzing the width of networks through optimization methods. We also shed light on two-layer networks with element-wise activation functions through experiments, indicating that neural networks have the potential to solve the number of equations equaling the number of parameters. As activation function, matrix exponential may provide less non-linearity as element-wise activation function do, but it may be possible to analyze based on the results in Lie theory. In the future, we will try to extend our method to multi-layer cases.

\section*{Acknowledgments}
This work has been supported by the CAS Project for Young Scientists in Basic Research [No. YSBR-034].

\bibliography{reference}
\bibliographystyle{iclr2024_conference}

\appendix

\end{document}